%% file: lrec-coling2024-example.tex
\documentclass[10pt, a4paper]{article}

\usepackage[]{lrec-coling2024} 

\usepackage{times}
\usepackage{latexsym}
\usepackage[T1]{fontenc}

\usepackage[utf8]{inputenc}

\usepackage{microtype}

\usepackage{inconsolata}
\usepackage{todonotes}

\usepackage{graphicx} 
\usepackage{multirow}
\usepackage{caption}
\usepackage{balance}
\usepackage{booktabs}
\usepackage{subcaption}

\usepackage{amsfonts}
\usepackage{amsmath}

\title{LANID: LLM-assisted New Intent Discovery}

\name{Lu Fan\textsuperscript{1}, Jiashu Pu\textsuperscript{2}, Rongsheng Zhang\textsuperscript{2}, and Xiao-Ming Wu\textsuperscript{1}$^{\ast}$ \thanks{*Corresponding author}}

\address{
Department of Computing, The Hong Kong Polytechnic University, Hong Kong S.A.R.\textsuperscript{1} \\ 
Fuxi AI Lab, NetEase Inc. China, Hang Zhou, China.\textsuperscript{2} \\
cslfan@comp.polyu.edu.hk, pujiashu@corp.netease.com, \\zhangrongsheng@corp.netease.com, xiao-ming.wu@polyu.edu.hk\\
}

\abstract{
Task-oriented Dialogue Systems (TODS) often face the challenge of encountering new intents. New Intent Discovery (NID) is a crucial task that aims to identify these novel intents while maintaining the capability to recognize existing ones. Previous efforts to adapt TODS to new intents have struggled with inadequate semantic representation or have depended on external knowledge, which is often not scalable or flexible.
Recently, Large Language Models (LLMs) have demonstrated strong zero-shot capabilities; however, their scale can be impractical for real-world applications that involve extensive queries. To address the limitations of existing NID methods by leveraging LLMs, we propose LANID, a framework that enhances the semantic representation of lightweight NID encoders with the guidance of LLMs.
Specifically, LANID employs the $K$-nearest neighbors and Density-Based Spatial Clustering of Applications with Noise (DBSCAN) algorithms to sample selective utterance pairs from the training set. It then queries an LLM to ascertain the relationships between these pairs. 
The data produced from this process is utilized to design a contrastive fine-tuning task, which is then used to train a small encoder with a contrastive triplet loss.
Our experimental results demonstrate the efficacy of the proposed method across three distinct NID datasets, surpassing strong baselines in both unsupervised and semi-supervised settings. Our code is available at \url{https://github.com/floatSDSDS/LANID}.
 \\ \newline  }

\begin{document}

\maketitleabstract

\section{Introduction}

Task-Oriented Dialogue Systems (TODS) are designed to assist users in accomplishing specific goals, such as booking flights or resolving technical issues. However, as these systems evolve, they inevitably encounter user intents that were unseen during initial training. New Intent Discovery (NID) aims to enable systems to autonomously identify novel intents while maintaining robust performance on known ones. Such capability is critical for deploying TODS in dynamic real-world environments. 
Previous approaches have utilized clustering algorithms~\citep{shi2018auto,perkins2019dialog,chatterjee2020intent} to group utterances with similar intents, or have implemented semi-supervised pre-training methods for representation learning.
These methods face significant limitations: many struggle to learn discriminative semantic representations for unseen intents~\citep{lin2020discovering,pu2022dialog, zhang2021discovering}, while others depend on the quality and quantity of external knowledge~\cite{zhang2021textoir, zhang2022new}, hindering scalability and adaptability.

Recent advancements in Large Language Models (LLMs)~\citep{openai2023gpt4} have demonstrated impressive zero-shot reasoning capabilities~\citep{heck2023chatgpt}, providing promising solutions for intent recognition. However, implementing these large-scale models in production environments is both complex and resource-intensive. This complexity raises potential privacy issues~\cite{kim2023propile}, and the significant computational demands can lead to increased latency, impacting the real-time performance essential for many applications. Consequently, there is a growing interest in exploring the use of LLMs to guide and optimize offline, lightweight NID models.

To leverage the capabilities of LLMs while ensuring lightweight performance and privacy for the NID task, we propose LANID (\textbf{L}LM-\textbf{A}ssisted \textbf{N}ew \textbf{I}ntent \textbf{D}iscovery), a novel framework that integrates the rich knowledge of LLMs into lightweight encoders using contrastive learning to enhance in-domain NID performance. 
Specifically, we propose to sampling pairs of utterances and using their relationships discriminated by LLMs as contrastive signals. As establishing relationships for all possible utterance pairs would incur excessively high costs, we propose two data sampling strategies to sample informative and representative utterance pairs. Both strategies leverage unsupervised algorithms to enhance data sampling. The initial strategy employs the concept of core points from DBSCAN~\cite{ester1996density} to effectively sample representative data. Conversely, the second strategy leverages the $K$-nearest neighbor algorithm~\cite{cover1967nearest} to enhance the probability that the sampled data will accurately represent class boundaries. The proposed selection strategies ensure that the chosen pairs are more representative and thereby enhancing the system's efficiency.

\par
Next, a powerful LLM is employed to determine the relationships between the selected utterance pairs. The obtained supervision signals are then used to construct a contrastive learning task. By optimizing a lightweight encoder with a triplet loss objective, LANID enhances the model’s ability to discern both known and new intents without compromising efficiency. The data sampling and teaching steps are conducted iteratively, and the training set can be incrementally updated with either LLM-annotated or manually annotated new data, ensuring the system's flexibility and scalability.

The main contributions of this paper are threefold: 
\begin{itemize}
    \item We propose a framework that distills LLM-derived semantic insights into small in-domain encoders with contrastive learning methods for the NID task.
    \item We introduce two data sampling strategies based on unsupervised algorithms to construct contrastive tasks more efficiently.
    \item Extensive experiments on three benchmark datasets demonstrate LANID’s superiority over existing baselines in both unsupervised and semi-supervised NID settings.
\end{itemize}

\input{src/figure/framework}

\section{Related Works}
We briefly review recent work on intent recognition, discovering new intents, and using LLMs to improve intent understanding in TODS.

\textbf{Intent Recognition.} To help TODS manage the ever-expanding range of user intents, intent recognition has been a long-standing and active research area. Current studies have looked into different settings and techniques, which can be grouped into three main categories: 
(1) Clustering-based approaches~\cite{shi2018auto,perkins2019dialog, chatterjee2020intent,pu2022dialog}, which group similar user queries to identify intents in an unsupervised manner;
(2)Few-shot intent recognition~\cite{zhang2021effectiveness, zhan2022closer, zhang2022fine}, which aims to enable models to quickly adapt to new scenarios with multiple new intents by creating different network structures or training frameworks; and 
(3) methods based on Pre-trained Language Models (PLMs)~\cite{haponchyk2018supervised, liu2019reconstructing, yan2020unknown, haponchyk2021supervised, ma2022effectiveness} believe that more general semantic representation can lead to better dialogue understanding.

\textbf{New Intent Discovery (NID).} To better suit real-world production needs, a more practical setting named NID has been introduced~\cite{zhang2021textoir,zhang2021discovering}. This setting effectively leverages known labels while also uncovering novel intent categories, thereby expanding the set of supported intents. Subsequently, further efforts have been made to expand on this line of research. 
Zhang et al.~\cite{zhang2022new} designed a specific contrastive loss to exploit self-supervisory signals in unlabeled data for clustering. Zhou et al.~\cite{zhou2023probabilistic} introduced a probabilistic framework for NID where intent assignments are treated as latent variables. An et al.~\cite{an2023new} proposed a robust pseudo-label training and source domain joint-training network to refine noisy pseudo labels and fully utilize prior knowledge.

\textbf{Large Language Models (LLMs).}
LLMs show great potential for advancing conversational AI~\citep{openai2023gpt4, heck2023chatgpt}; however, their application in NID is still not well explored. While LLMs excel at detecting emerging intents, they require extensive computational resources and pose privacy concerns. As mentioned above, existing NID methods typically rely on smaller semantic encoders, such as BERT~\cite{devlin2018bert}. Recently, Song et al.~\cite{song2023large} reveal ChatGPT's strengths in zero-shot settings but also pointed out its limitations compared to fine-tuned models. 
In this paper, we propose leveraging the semantic understanding capabilities of LLMs to generate auxiliary labels for contrastive training.

\section{Proposed Method: LANID}

\textbf{Problem Formulation.}
To better align with the practical need for continuous training to recognize new intents, we follow the approach of prior research~\citep{zhang2021discovering} and evaluate our method in both unsupervised and semi-supervised evaluation settings. 
We represent an utterance and its intent label as $x$ and $y$, respectively. $y$ can either belong to the set of unknown intents $\mathcal{C}_u$ or the set of known intents $\mathcal{C}_k$. 
The training set, validation set, and test set are denoted as $D_{train}$, $D_{val}$ and $D_{test}$. These three sets share the same distribution of intent labels. 
In the unsupervised setting, $D_{train}$ does not contain any annotated utterances, meaning that no intents are known or labeled. In the semi-supervised setting, $D_{train}$ comprises both a labeled dataset, $D_{labeled}=\{(x_i, y_i)|y_i \in \mathcal{C}_k\}$ and an unlabeled dataset $D_{unlabeled}$. The intent labels of utterances in $D_{unlabeled}$ may belong to either $\mathcal{C}_k$ or $\mathcal{C}_u$. In both settings, the goal is to group utterances from $D_{test}$ into clusters, with each cluster representing a distinct intent. 

\input{src/table/parameters}

\input{src/table/rst_unsupervised}

\paragraph{Overview.} The proposed LANID framework is illustrated in Figure~\ref{fig:method}. LANID utilizes contrastive learning to distill knowledge from LLMs, thereby facilitating the training of a lightweight encoder for novel intent detection. The contrastive learning process consists of three main steps: 1) sampling selective utterance pairs from \(D_{\text{train}}\) to form the contrastive tasks based on clustering algorithms, 2) determining the relationships between these utterance pairs using a powerful LLM that provides pseudo-labeling, and 3) fine-tuning the encoder with the LLM's output to distill the knowledge from the powerful LLM. This three-step training procedure is iteratively repeated until either a predefined condition set by the user is satisfied or convergence is attained.
Finally, the utterances in $D_{\text{test}}$ are encoded using the optimized encoder and clustered with the $k$-means algorithm, consistent with previous attempts in NID~\cite{zhang2021discovering, zhang2022new}.

\input{src/table/rst_semi}

\subsection{Selecting Utterance Pairs} \label{sec:sampler}
We propose injecting knowledge encoded in LLMs into the small model through contrastive learning. In the learning process, the LLM acts like a teacher, while the small model acts like a student. One of the most essential things in the teaching stage is designing contrastive tasks. In this work, we propose sampling utterance pairs and asking the LLM to determine the relation between each one. The contrastive tasks are formed with utterance pairs and their pseudo-relation label annotated by LLMs.
It is impractical to establish relationships for all possible utterance pairs. Therefore, we propose two selective sampling strategies for utterance pair sampling, which will selectively sample representative and informative data. Next, we will illustrate the two proposed sampling strategies.

\noindent \paragraph{Selection based on $K$-Nearest Neighbors.} 
Starting locally, we first find those utterances that are close to each other (based on the original representation) and determine whether the distribution among them is reasonable. In each iteration, we randomly sample $p\%$ utterances from $D_{train}$. Then, for each sampled utterance $x_i$, we search for its top-$K$ nearest neighbors $\mathcal{N}_i^{Near}$ using the Euclidean distance, and we uniformly sample $n_k (n_k < |\mathcal{N}_i^{Near}|)$ utterances from $\mathcal{N}_i^{Near}$. We denote the nearest-neighbor set for $x_i$ as $\mathcal{M}_i^{Near} = \{(x_i, x_j) | x_j \in \mathcal{N}_i^{Near}\}$, where $|\mathcal{M}_i^{Near}|=n_k$.

\noindent \paragraph{Selection based on Global Density.} 
It is difficult to divide a data set into exactly a few categories, and there will always be some outliers. Also, the distribution of semantics is usually not uniform, and there are high and low densities of different semantic clusters. We propose a DBSCAN-based~\citep{ester1996density} sampling approach to reflect the relationship between globally high and low-density regions of semantics. Concretely, we conduct DBSCAN clustering on $D_{train}$ and obtain a set of core points $\mathbf{x}_{c}$ and a set of non-core points $\mathbf{x}_{nc}$. Then, we randomly sample a subset  $\mathbf{x}_{nc}^{'}$ from $\mathbf{x}_{nc}$. For each utterance $x_i$  in  $\mathbf{x}_{nc}^{'}$, we search for its $m$ nearest neighbors in $\mathbf{x}_{c}$, forming a global density set as $\mathcal{M}_i^{Den} = \{(x_i, x_j) | x_j \in \mathcal{N}_i^{Core}\}$, where $\mathcal{N}_i^{Core}$ is the set consisting of the nearest points to $x_i$ in $\mathbf{x}_{c}$.

\subsection{LLM Manager}

The LLM manager is another major module in LANID. It constructs prompts using sampled data and parses the responses from LLMs. 
We design prompts with three components~\citep{pan2023preliminary}: schema, regulations, and sentence input. The schema component is intended to guide LLMs to produce responses that meet our desired criteria. To identify the optimal schema for each dataset, several schemas were manually crafted and evaluated based on their performance on $D_{labeled}$. The regulations component constrains the format of the LLM's responses. For simplicity, we uniformly use the phrase "just answer yes or no." The sentence input component consists of utterance pairs sampled as detailed in Section~\ref{sec:sampler}. Finally, we predict $r(i, j)=1$ for a data pair $(i, j)$ if 'yes' appears in the LLM's corresponding response; otherwise, $r(i, j)=0$.

\subsection{Training and Optimization}

To optimize the representation of the text encoder on domain data, we collect pairs of positive samples from $\mathcal{M}_i^{Near}$ and/or $\mathcal{M}_i^{Core}$, with their relationships $r(i, j)$ determined by the LLM manager. For each positive sample pair $\{(x_i, x_j)\}$, we randomly sample $k_n$ utterances from $D_{\text{train}}$ as negative samples to better represent the distribution of the entire dataset, assuming the dataset distribution is not extreme. This approach forms a dataset $D_f={(x_i, p_i, n_i)}$ of triplets. We then fine-tune the model using a triplet margin loss defined as:
\begin{equation}
\begin{aligned}
    \mathcal{L}(x_i, p_i, n_i) = max(
    d(x_i, p_i)-d(x_i, n_i)+ mgn, 0),
\end{aligned}
\end{equation}
where $x_i$ serves as the anchor point, mgn. represents a hyperparameter, and $p_i$ and $n_i$ are its positive and negative samples, respectively. The function $d(x_i, y_j)=\Vert x_i - y_j \Vert$ represents the distance, and the margin is a hyperparameter that determines the minimum desired difference between $d(x_i, p_i)$ and $d(x_i, n_i)$.

As training progresses, the quality of the text encoder's representation improves, leading to enhanced sampling outcomes. In practice, the process of selecting utterance pairs and querying the LLM manager recurs every $T$ epochs, with the fine-tuning dataset $D_f$ and the text encoder being incrementally updated during this iterative process.

Finally, upon completing the training stage, we apply the optimized encoder on the test set $D_{\text{test}}$ and employ the $k$-means algorithm to do clustering. Utterances that fall within the same cluster are considered to share the same intent.

\section{Experiment}

\subsection{Experimental Details}

\noindent
We evaluate LANID using three intent recognition benchmarks. The BANKING dataset \citep{casanueva2020efficient} includes 13,083 utterances distributed across 77 intents within the banking domain. The StackOverflow dataset \citep{xu2015short} comprises 20,000 queries collected from an online question-answering platform\footnote{https://stackoverflow.com/}, categorized into 20 different categories. The M-CID dataset \citep{arora2020cross} consists of 1,745 utterances associated with 16 intents specifically related to COVID-19 services.

\noindent
\paragraph{Experimental Setup.} 
Our proposed method is evaluated under both unsupervised and semi-supervised settings. We use three clustering evaluation metrics: normalized mutual information (NMI) \citep{estevez2009normalized}, adjusted Rand index (ARI) \citep{yeung2001details}, and accuracy (ACC).

\noindent
\paragraph{Baselines.} 
We compare LANID with both unsupervised and semi-supervised NID state-of-the-art (SOTAs). The unsupervised NID SOTAs include the SAE series \citep{xie2016unsupervised}, MTP, and CLNN \citep{zhang2022new}. The semi-supervised baselines include BERT-KCL \citep{hsu2019multi}, DAC \citep{zhang2021discovering}, MTP, and CLNN \citep{zhang2022new}.

\noindent 
\paragraph{Implementation Details.} 
We use the default settings of CLNN \citep{zhang2022new} and continue to train the model pretrained by MTP-CLNN as a post-finetuning stage. Further training can also be conducted on other NID baselines. For LLMs, we use the \textit{gpt-3.5-turbo}\footnote{https://platform.openai.com/docs/models/gpt-3-5} model. The hyperparameters of LANID as shown in Table~\ref{tb:param} are selected based on performance on the validation set.

\subsection{Result Analysis}

Tables~\ref{tb:unsupervised} and \ref{tb:semi} summarize the performance of LANID compared to state-of-the-art (SOTA) methods in both unsupervised and semi-supervised settings across three intent recognition benchmarks. 
The results reveal several key observations: (1) LANID and its variants excel in unsupervised learning, beating other baselines due to effective LLM labeling that compensates for missing supervised signals. (2) LANID consistently surpasses baselines in most cases, proving its effectiveness. (3) Although the combination of two neighborhood sampling strategies generally works well, relying solely on the DBSCAN-based sampling strategy can sometimes hinder performance. This is because the LLM is constrained to make binary judgments and retain only positive pairs. For many outliers, their nearest cores may not express the same intent, thereby reducing the size of $D_f$ and leading to overfitting problems.

\section{Conclusion}

   In this work, we introduce LANID, a framework designed to tackle the challenge of novel intent discovery in TODS by LLMs to train lightweight encoders through contrastive learning. This approach enhances the system's capability to identify new user intents while maintaining efficiency, scalability, and privacy. By employing effective data sampling techniques and harnessing LLM knowledge, LANID demonstrates superior performance compared to existing methods, as evidenced by extensive experiments on benchmark datasets. Overall, the proposed framework sets the stage for more adaptive and responsive TODS that can effectively operate in dynamic real-world environments.

\section*{Acknowledgments}
We thank the anonymous reviewers for their valuable feedback. This research was partially supported by the grant of HK ITF ITS/359/21FP.

\nocite{*}
\section{Bibliographical References}\label{sec:reference}
\bibliographystyle{lrec-coling2024-natbib}
\bibliography{lrec-coling2024-example}

\bibliographystylelanguageresource{lrec-coling2024-natbib}
\bibliographylanguageresource{languageresource}

\end{document}

%% file: src/figure/framework.tex
\begin{figure*}
    \centering
    \includegraphics[width=0.8\linewidth]{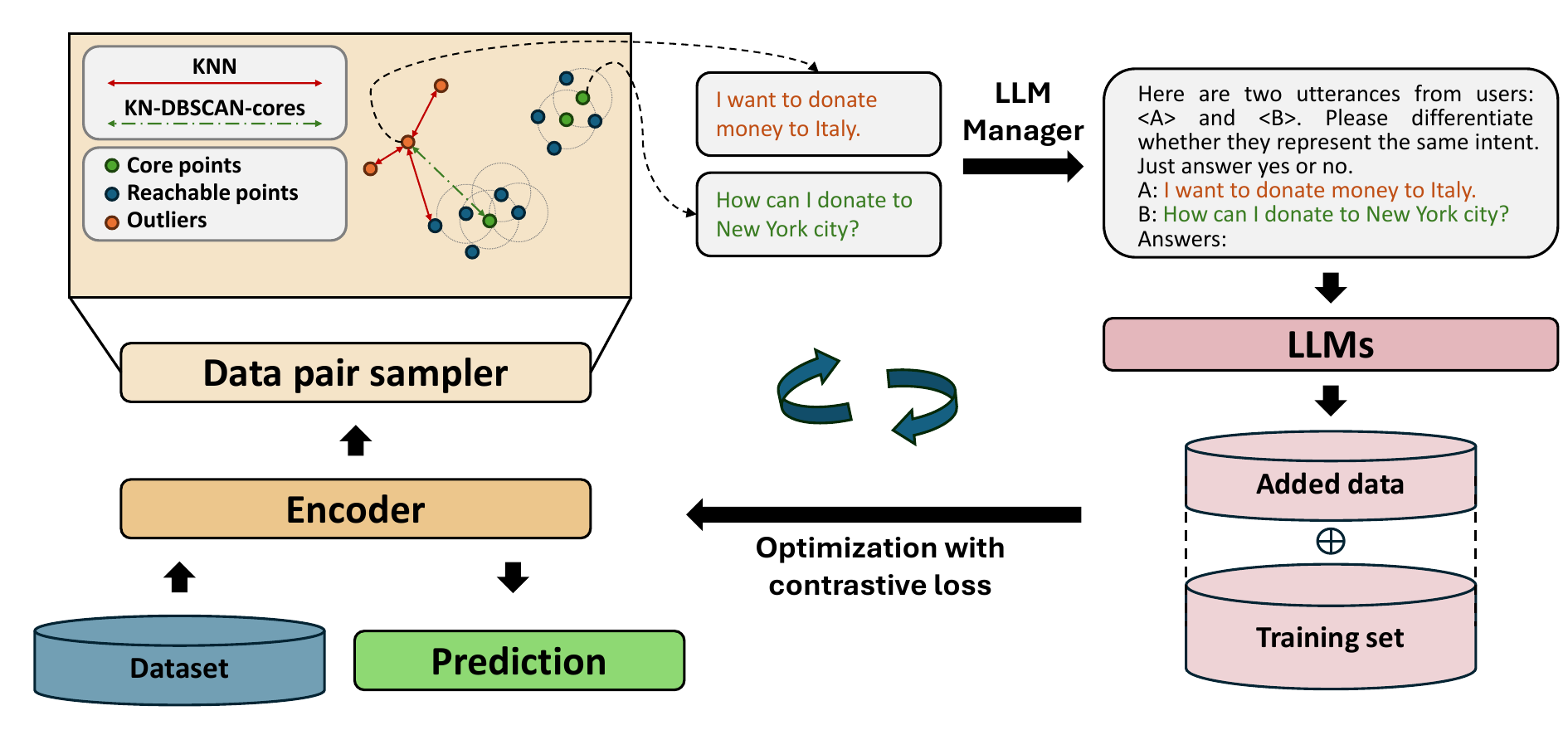}
    \caption{Illustration of the proposed LANID framework. First, utterances are encoded by our target light encoder, and informative utterance pairs are sampled from $D_{train}$ using the proposed data pair sampler. Next, the LLM manager inputs the sampled data pairs into the LLM and uses it to annotate the relationship for each utterance pair. Contrastive tasks are then constructed to optimize the target encoder. 
    These three steps are repeated iteratively until convergence is reached. Once this is achieved, we apply the trained encoder to the test set, $D_{test}$, and utilize the $k$-means algorithm to cluster the encoded utterances. Utterances within the same cluster are then considered to share the same intent.}
    \label{fig:method}
\end{figure*}

%% file: src/table/parameters.tex
\begin{table*}[]
\small
\centering
\caption{Hyper-parameter Settings: MinPts denotes the minimum number of points needed within a specified radius (epsilon) to form a dense region in DBSCAN.}
\scalebox{0.75}{
\begin{tabular}{lllllllll}
\hline
              & $K$  & $p$    & $n_k$ & $m$ & $k_n$ & $T$ & \#Epoch & MinPts \\ \hline
Banking       & 50 & 0.1  & 2    & 5 & 2    & 3 & 10      & 4      \\
Stackoverflow & 50 & 0.05 & 2    & 8 & 2    & 2 & 10      & 4      \\
M-CID         & 50 & 0.2  & 2    & 5 & 2    & 3 & 20      & 4      \\ \hline
\end{tabular}
}
\label{tb:param}
\end{table*}

%% file: src/table/rst_unsupervised.tex
\begin{table*}[]
\centering
\caption{
Performance on Unsupervised NID: Best results for each dataset are highlighted in bold. Results are shown for three LANID variants: LANID-Near (KNN-based sampling), LANID-DBSCAN (DBSCAN sampling), and LANID (combination of both strategies)}
\scalebox{0.7}{
\begin{tabular}{cllllllllll}
\toprule
\hline
\multicolumn{1}{l}{\textbf{}}  & \textbf{}        & \multicolumn{3}{c}{\textbf{Banking}}       & \multicolumn{3}{c}{\textbf{StackOverflow}} & \multicolumn{3}{c}{\textbf{M-CID}}         \\
\multicolumn{1}{l}{\textbf{}}  & \textbf{Methods} & \textbf{NMI} & \textbf{ARI} & \textbf{ACC} & \textbf{NMI} & \textbf{ARI} & \textbf{ACC} & \textbf{NMI} & \textbf{ARI} & \textbf{ACC} \\ \hline
\multirow{11}{*}{} & SAE-KM           & 60.12          & 24.00          & 37.38          & 48.72          & 23.36          & 37.16          & 51.03          & 43.51          & 52.95          \\
                               & SAE-DEC          & 62.92          & 25.68          & 39.35          & 61.32          & 21.17          & 57.09          & 50.69          & 44.52          & 53.07          \\
                               & SAE-DCN          & 62.94          & 25.69          & 39.36          & 61.34          & 34.98          & 57.09          & 50.69          & 44.52          & 53.07          \\
                               & MTP              & 77.25          & 47.80          & 59.12          & 61.35          & 45.77          & 61.90          & 70.53          & 45.76          & 64.76          \\
                               & MTP-CLNN         & 82.15          & 57.68          & 66.88          & 75.20          & 63.13          & 79.20          & 80.03          & 67.39          & 79.94          \\ \cline{2-11} 
                               & LANID-Near     & 83.44          & 58.28          & 66.75          & 79.56          & 66.67          & 83.40          & 80.80          & 69.86          & 81.38          \\
                               & LANID-DBSCAN   & 83.21          & 58.02          & 65.78          & 81.25          & 72.86          & 85.30          & 80.41          & 68.10          & 79.08          \\
                               & LANID          & \textbf{84.12} & \textbf{60.40} & \textbf{70.58} & \textbf{81.25} & \textbf{72.96} & \textbf{86.60} & \textbf{82.64} & \textbf{71.36} & \textbf{82.52} \\ \hline \bottomrule
\end{tabular}
}
\label{tb:unsupervised}
\end{table*}

%% file: src/table/rst_semi.tex
\begin{table*}[]
\centering
\caption{
Performance on Semi-Supervised NID with Varying Known Class Ratios: Best results for each dataset are highlighted in bold. The Known Class Ratio (KCR) is defined as 
$\frac{|\mathcal{C}_k|}{(|\mathcal{C}_k|+\mathcal{C}_u)}$. A 10\% subset of each known class was randomly sampled to create $D_{labeled}$. Results are shown for the three LANID variants as described in Table~\ref{tb:unsupervised}.
}
\scalebox{0.75}{
\begin{tabular}{cllllllllll}
\toprule
\hline
\multicolumn{1}{l}{\textbf{}} & \textbf{}        & \multicolumn{3}{c}{\textbf{Banking}}             & \multicolumn{3}{c}{\textbf{StackOverflow}}       & \multicolumn{3}{c}{\textbf{M-CID}}               \\
\multicolumn{1}{l}{\textbf{}} & \textbf{Methods} & \textbf{NMI}   & \textbf{ARI}   & \textbf{ACC}   & \textbf{NMI}   & \textbf{ARI}   & \textbf{ACC}   & \textbf{NMI}   & \textbf{ARI}   & \textbf{ACC}   \\ \hline
\multirow{9}{*}{KCR-25\%}     & BERT-KCL         & 53.85          & 20.07          & 28.79          & 35.47          & 16.80          & 32.88          & 29.35          & 11.58          & 24.76          \\
                              & DAC              & 69.85          & 37.16          & 49.67          & 53.97          & 36.46          & 53.96          & 49.83          & 27.21          & 43.72          \\
                              & MTP              & 79.17          & 50.83          & 62.05          & 74.86          & 62.27          & 77.20          & 70.53          & 45.76          & 64.76          \\
                              & MTP-CLNN         & 83.88          & 60.76          & 70.91          & 78.38          & 65.80          & 80.10          & 78.30          & 65.32          & 78.30          \\ \cline{2-11} 
                              & LANID-Near     & 85.28          & 63.48          & \textbf{72.47} & \textbf{80.83} & \textbf{65.86} & \textbf{83.30} & 81.91          & 70.30          & 81.09          \\
                              & LANID-DBSCAN   & 84.74          & 62.22          & 70.13          & 74.74          & 60.54          & 73.70          & 80.04          & 69.69          & 83.09          \\ 
                              & LANID          & \textbf{85.51} & \textbf{64.23} & 71.40          & 79.55          & 63.23          & 81.80          & \textbf{85.11} & \textbf{75.66} & \textbf{86.82} \\ \hline
\multicolumn{1}{l}{\textbf{}} & \textbf{}        & \multicolumn{3}{c}{\textbf{Banking}}             & \multicolumn{3}{c}{\textbf{StackOverflow}}       & \multicolumn{3}{c}{\textbf{M-CID}}               \\
\multicolumn{1}{l}{\textbf{}} & \textbf{Methods} & \textbf{NMI}   & \textbf{ARI}   & \textbf{ACC}   & \textbf{NMI}   & \textbf{ARI}   & \textbf{ACC}   & \textbf{NMI}   & \textbf{ARI}   & \textbf{ACC}   \\ \hline
\multirow{9}{*}{KCR-50\%}     & BERT-KCL         & 62.86          & 30.16          & 40.81          & 57.63          & 41.90          & 56.58          & 42.48          & 22.83          & 38.11          \\
                              & DAC              & 76.41          & 47.28          & 59.32          & 70.78          & 56.44          & 73.76          & 63.27          & 43.52          & 57.19          \\
                              & MTP              & 82.12          & 56.43          & 67.34          & 76.58          & 65.55          & 82.50          & 70.53          & 45.76          & 64.76          \\
                              & MTP-CLNN         & 86.42          & 66.66          & 74.97          & 81.41          & \textbf{72.15} & \textbf{86.00} & 79.34          & 66.18          & 78.80          \\ \cline{2-11} 
                              & LANID-Near     & \textbf{86.83} & \textbf{67.41} & \textbf{76.10} & 81.62 & 64.32          & 81.30          & 81.20          & 69.54          & 81.95          \\
                              & LANID-DBSCAN   & 85.62          & 64.35          & 72.44          & 81.19          & 65.75          & 81.40          & 79.16          & 67.85          & 80.80          \\
                              & LANID          & 86.31          & 66.53          & 75.49          & \textbf{82.07}          & 70.51          & 83.00          & \textbf{81.58} & \textbf{70.66} & \textbf{82.81} \\ \hline
\multicolumn{1}{l}{\textbf{}} & \textbf{}        & \multicolumn{3}{c}{\textbf{Banking}}             & \multicolumn{3}{c}{\textbf{StackOverflow}}       & \multicolumn{3}{c}{\textbf{M-CID}}               \\
\multicolumn{1}{l}{\textbf{}} & \textbf{Methods} & \textbf{NMI}   & \textbf{ARI}   & \textbf{ACC}   & \textbf{NMI}   & \textbf{ARI}   & \textbf{ACC}   & \textbf{NMI}   & \textbf{ARI}   & \textbf{ACC}   \\ \hline
\multirow{9}{*}{KCR-75\%}     & BERT-KCL         & 72.18          & 44.29          & 58.70          & 70.38          & 57.98          & 71.50          & 54.22          & 34.60          & 52.15          \\
                              & DAC              & 79.99          & 54.57          & 65.87          & 75.31          & 60.02          & 78.84          & 71.41          & 54.22          & 69.11          \\
                              & MTP              & 84.61          & 63.23          & 72.76          & 80.41          & 70.01          & 81.10          & 77.90          & 64.57          & 77.65          \\
                              & MTP-CLNN         & 87.24          & 68.77          & 77.14          & 80.99          & 72.14          & 85.80          & 80.12          & 67.40          & 79.37          \\  \cline{2-11} 
                              & LANID-Near     & 87.59          & \textbf{70.13} & \textbf{78.51} & 82.14          & 73.05          & 85.80          & 81.13          & 69.75          & \textbf{83.09} \\
                              & LANID-DBSCAN   & 86.79          & 67.18          & 74.35          & \textbf{83.74} & \textbf{76.45} & \textbf{88.30} & 80.65          & 70.24          & 82.52          \\
                              & LANID          & \textbf{87.64} & 68.89          & 76.56          & 82.80          & 74.33          & 87.50          & \textbf{82.16} & \textbf{70.56} & 82.23          \\ \hline \bottomrule
\end{tabular}
}
\label{tb:semi}
\end{table*}